%% file: tacl2021v1-template.tex
\documentclass[11pt,a4paper]{article}

\usepackage{inconsolata}
\usepackage[T1]{fontenc}
\usepackage{times,latexsym}
\usepackage{amsmath}
\usepackage{amsfonts}
\usepackage{amsthm}
\usepackage{graphicx}
\usepackage[dvipsnames,table]{xcolor}
\usepackage{enumitem}
\usepackage{url}
\usepackage{comment}
\usepackage{lipsum}
\usepackage[edges]{forest}
\usepackage{microtype}
\usepackage{booktabs}
\usepackage{tcolorbox}
\usepackage{subcaption}
\usepackage{tikz}
\usetikzlibrary{positioning}

\usepackage[acceptedWithA]{tacl2021v1}

\usepackage{xspace,mfirstuc,tabulary}

\newif\iftaclinstructions
\taclinstructionsfalse 
\iftaclinstructions
\renewcommand{\confidential}{}
\renewcommand{\anonsubtext}{(No author info supplied here, for consistency with
TACL-submission anonymization requirements)}
\newcommand{\instr}
\fi

\iftaclpubformat 

\else

\fi

\input{macros.tex}

\title{
  Multilinguality at the Edge:\\Developing Language Models for the Global South
}

\input{authors.tex}

\date{}

\defcitealias{omnilingualasrteam2025omnilingualasropensourcemultilingual}{Omni.\ Team}

\begin{document}
\maketitle
\begin{abstract}
  Where and how language models (LMs) are deployed determines who can benefit from them.
  However, there are several challenges that prevent effective deployment of LMs in non-English-speaking and hardware-constrained communities in the Global South.
  We call this challenge the \textit{last mile}: the intersection of multilinguality and edge deployment, where the goals are aligned but the technical requirements often compete.
  Studying these two fields together is both a \textit{need}, as linguistically diverse communities often face the most severe infrastructure constraints, and an \textit{opportunity}, as edge and multilingual NLP research remain largely siloed.
  To understand the state of the art and the challenges of combining the two areas, we survey \numpapers{} papers that tackle this problem across the language modelling pipeline, from data collection to development and deployment.
  We also discuss open questions and provide actionable recommendations for different stakeholders in the NLP ecosystem.
  Finally, we hope that this work contributes to the development of inclusive and equitable language technologies.
\end{abstract}

\section{Introduction}

Language models (LMs) have made remarkable progress in recent years and are increasingly integrated into everyday work and life \citep{microsoft2026aidiffusion}.
This progress is evident in both multilingual coverage, with models now speaking hundreds of languages \citep{ustun-etal-2024-aya,gemmateam2025gemma3technicalreport},
and in efficiency, with capable models now small enough to deploy on smartphones and other edge devices \citep[\textit{inter alia}]{treviso-etal-2023-efficient}.

Yet despite these advances, a wide gap persists in who can actually deploy and use these models, particularly between the Global North and the Global South \citep{joshi-etal-2020-state,khan2024ailics,pava2025mind}.
We can point to several factors that contribute to this disparity.
Most LMs are not trained on the long tail of the world's 7,000 languages and thus lack the performance necessary for downstream applications \citep{occhini2026artificialintelligencecreatingnew}.
In addition, infrastructure constraints such as limited connectivity and low smartphone penetration, especially in the Global South, further restrict the use of both cloud-based and on-device models \citep{GSMA2025MobileEconomy,itu-2025-facts}.
As LMs become central to economic productivity, these barriers threaten to widen the divide further.

\begin{figure*}[t]
  \centering
  \begin{tikzpicture}
    \node[anchor=south west, inner sep=0] (image)
    {\includegraphics[width=0.90\linewidth, trim={1cm 1.2cm 1cm 0}]{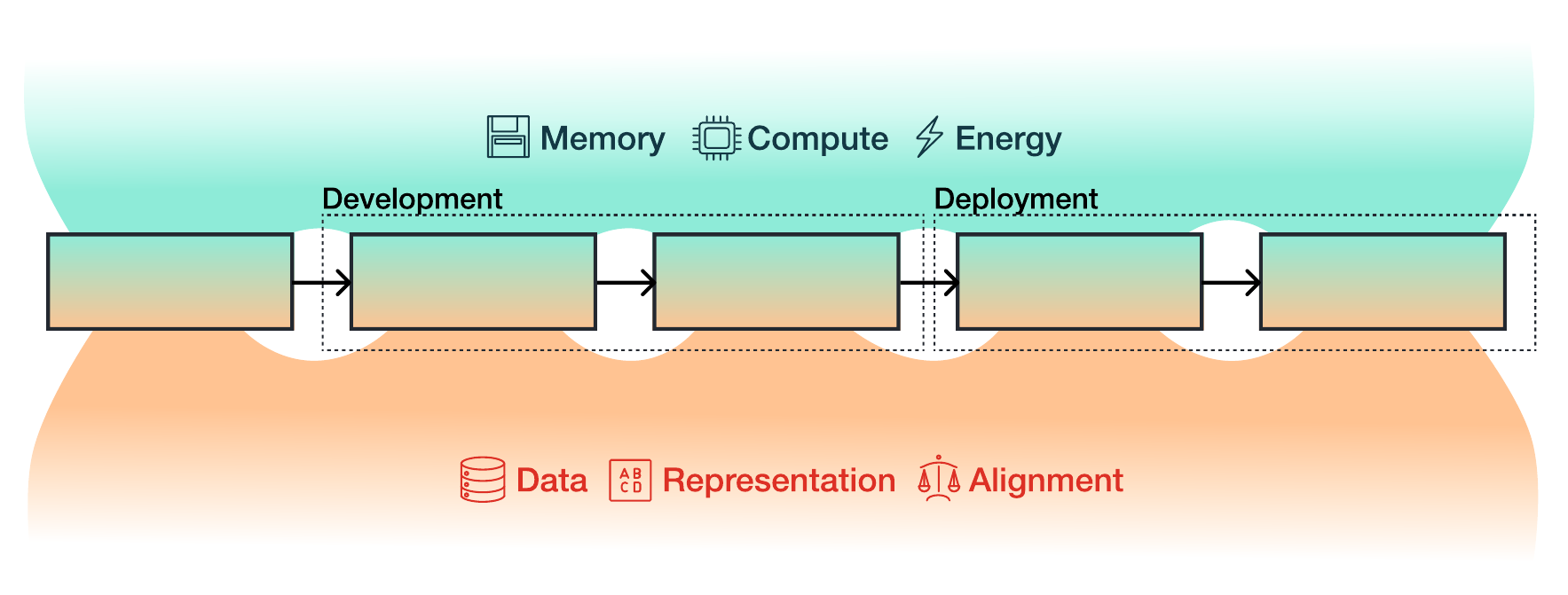}};
    \begin{scope}[x={(image.south east)}, y={(image.north west)}]
      \node[anchor=center, font=\fontfamily{phv}\selectfont\large\bfseries, text=cdarkblue] at (0.48, 0.86)
      {Requirements of the Edge (\S\ref{sec:edge_constraints})};
      \node[anchor=center, font=\fontfamily{phv}\selectfont\large\bfseries, text=cdarkcrest] at (0.52, 0.20)
      {Requirements for Multilingual Capability (\S\ref{sec:multilingual_constraints})};
      \node[anchor=center, font=\fontfamily{phv}\selectfont\small\bfseries] at (0.08, 0.495)
      {\begin{tabular}{c}Data Collect.\end{tabular}};
      \node[anchor=center, font=\fontfamily{phv}\selectfont\small] at (0.08, 0.425)
      {(\S\ref{sec:data_collection})};
      \node[anchor=center, font=\fontfamily{phv}\selectfont\small\bfseries] at (0.29, 0.495)
      {Pretraining};
      \node[anchor=center, font=\fontfamily{phv}\selectfont\small] at (0.29, 0.425)
      {(\S\ref{sec:pretraining})};
      \node[anchor=center, font=\fontfamily{phv}\selectfont\small\bfseries] at (0.495, 0.495)
      {Post-training};
      \node[anchor=center, font=\fontfamily{phv}\selectfont\small] at (0.495, 0.425)
      {(\S\ref{sec:post_training})};
      \node[anchor=center, font=\fontfamily{phv}\selectfont\small\bfseries] at (0.70, 0.495)
      {Inference};
      \node[anchor=center, font=\fontfamily{phv}\selectfont\small] at (0.70, 0.425)
      {(\S\ref{sec:inference})};
      \node[anchor=center, font=\fontfamily{phv}\selectfont\small\bfseries] at (0.91, 0.495)
      {Evaluation};
      \node[anchor=center, font=\fontfamily{phv}\selectfont\small] at (0.91, 0.425)
      {(\S\ref{sec:evaluation})};
    \end{scope}
  \end{tikzpicture}
  \caption{
    \textbf{Competing requirements at each step of the language modelling pipeline.}
    Edge LM deployment imposes constraints on memory, compute, and energy \citep{treviso-etal-2023-efficient,zheng2025reviewedgeLLM}, which can conflict with the requirements for building capable multilingual models \citep{longpre2025atlasadaptivetransferscaling,yong-etal-2025-state}.
    This survey examines how these competing demands manifest across the pipeline and reviews techniques that address them.
  }
  \label{fig:lm_pipeline}
\end{figure*}
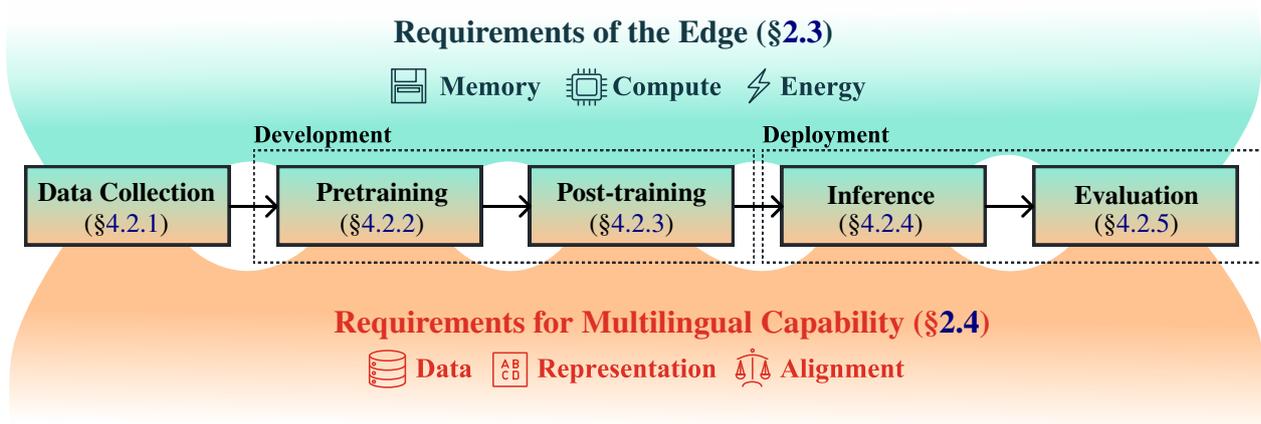

Reaching these communities through language technology requires a holistic understanding of their contexts, specifically on whether LMs can understand their language and run on their hardware.
Tackling the problems of efficiency or multilinguality in isolation has merit \citep{ruder-etal-2022-square}, but it risks not addressing the actual needs of the communities left behind this widening gap.
For example, research on model compression has largely focused on English, with several studies observing degradation in multilingual performance as models are made smaller \citep{ogueji-etal-2022-intriguing,mohammadshahi-etal-2022-compressed,marchisio-etal-2024-quantization}.
Similarly, multilingual scaling laws \citep{longpre2025atlasadaptivetransferscaling,he-etal-2025-scaling} reveal a power-law relationship between multilingual performance and model scale, suggesting that smaller, more deployable models struggle with low-resource languages.
Approaching these lines of work in tandem is important, as \citet{ahia-etal-2021-low-resource} argue that language communities face constraints not only in terms of data, but also of compute (the ``low-resource double bind'').
This view is echoed by \citet{nigatu-etal-2024-zenos}, who show that ``low-resourceness'' pertains to technology as much as data.
Building on these insights, we then ask: \textit{``(1) what challenges and methods shape multilingual edge LM development, and (2) what opportunities exist for reaching the communities that need these technologies the most?''}

We call this challenge the \textit{last mile}, which refers to the intersection of multilinguality and edge deployment,
reaching communities that need both a model that understands their language and the efficiency to run on their available hardware devices.
This challenge is non-trivial, as we show that while the goals of edge deployment and multilinguality are aligned, their technical requirements often compete.
In addition, decisions made during development can directly affect whether a model is deployable on constrained hardware.

In this work, we define the challenge of the \textit{last mile} (\S\ref{sec:multilinguality_at_the_edge}) and survey \numpapers{} papers that address this problem at different stages of the language modelling pipeline (\S\ref{sec:lm_pipeline}), characterizing the state of the art and the tensions between the competing requirements of edge deployment and multilinguality.
For each stage, we describe how the constraints of edge deployment (\ref{sec:edge_constraints}) and the requirements of multilinguality (\ref{sec:multilingual_constraints}) can impose competing demands, and how existing work resolves the gap (\S\ref{sec:survey}).
In addition, to stimulate new work at this intersection of multilingual and edge NLP, we discuss opportunities and provide actionable recommendations for different stakeholders (\S\ref{sec:discussion}).
Finally, we hope that this work contributes to the development of equitable language technologies.

\section{Multilinguality at the Edge}
\label{sec:multilinguality_at_the_edge}

\subsection{Why Study the Two Fields Together?}
\label{sec:motivation}

While Edge LM and Multilinguality have each been studied in isolation,
we observe trends suggesting that they should be studied together.

\begin{figure}[t]
  \centering
  \includegraphics[width=0.82\linewidth, trim={1.2cm 0.8cm 1cm 0}]{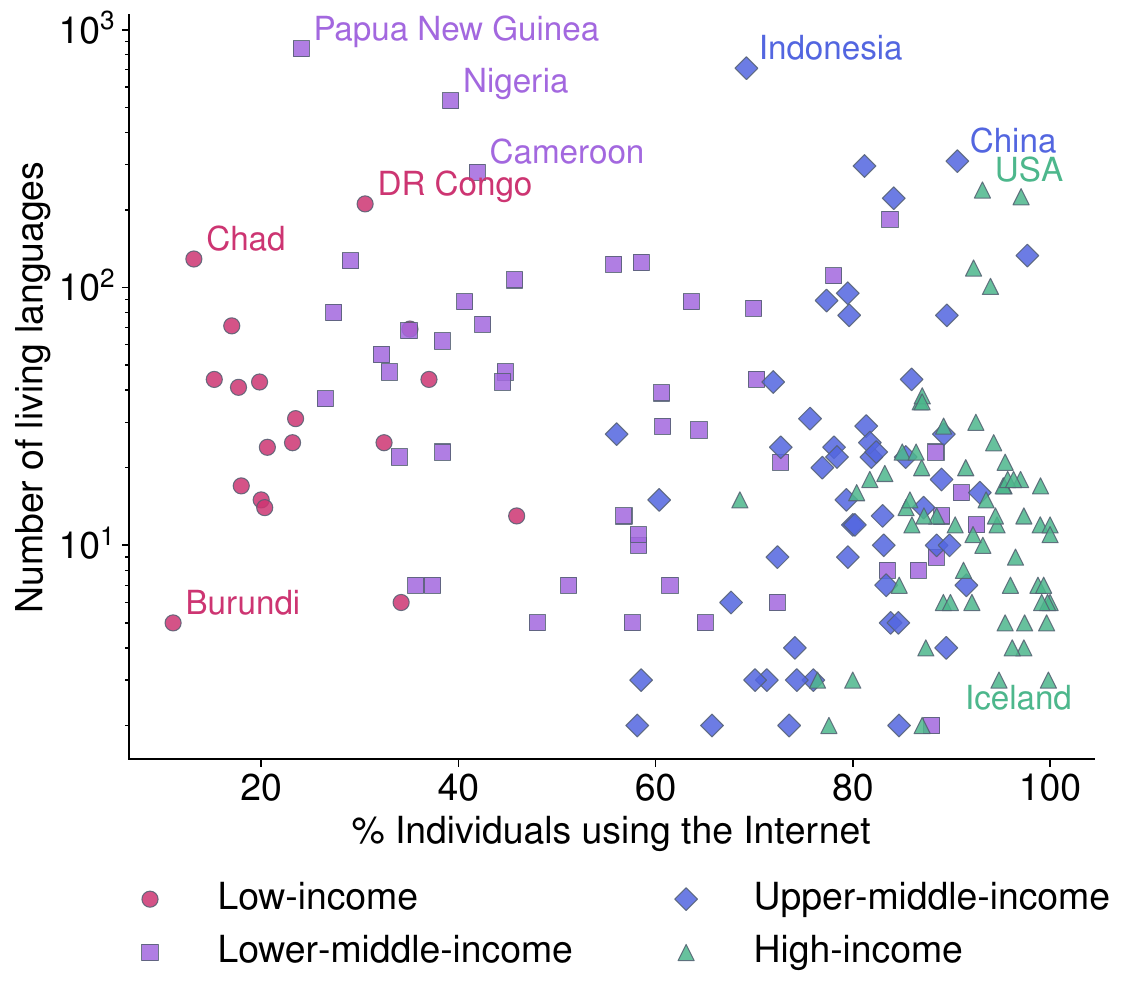}
  \caption{
    \textbf{Countries with high linguistic diversity have the most limited network connectivity (upper left).}
    Internet penetration is sourced from \citet{itu-2025-facts},
    number of living languages ($\log$-scale) from the Ethnologue \citep{ethnologue2025},
    and income groups from the \citet{worldbank2025income}.
  }
  \label{fig:infra_lingdiv_ict}
\end{figure}

\paragraph{Need: Multilingual communities face infrastructure constraints}
The first trend is that communities with the highest linguistic diversity often face the most severe infrastructure constraints. 
As shown in \autoref{fig:infra_lingdiv_ict}, countries such as Papua New Guinea, Nigeria, and Chad are linguistically diverse yet among the least active on the Internet.
This compounds their disadvantage in the LM pipeline: most pretraining data are derived from web-scale text, which these communities lack due to limited Internet activity, and
the dominant paradigm for serving LMs relies on cloud infrastructure (i.e., requests are sent over a network). 
This leads to LMs that are both unable to serve their languages and undeployable in their specific contexts.
In addition, entry-level smartphones cost up to 48\% of monthly income for the poorest quintile in these regions, meaning the devices that do exist are mostly low-end \cite{GSMA2025MobileEconomy}.
Of the 21 countries with more than 100 living languages, 12 are classified as low or lower-middle income \citep{worldbank2025income}, showing a \textit{need} for models that can run entirely on-device.

\begin{figure}[t]
  \centering
  \includegraphics[width=\linewidth, trim={0cm 0.5cm 0cm 0}]{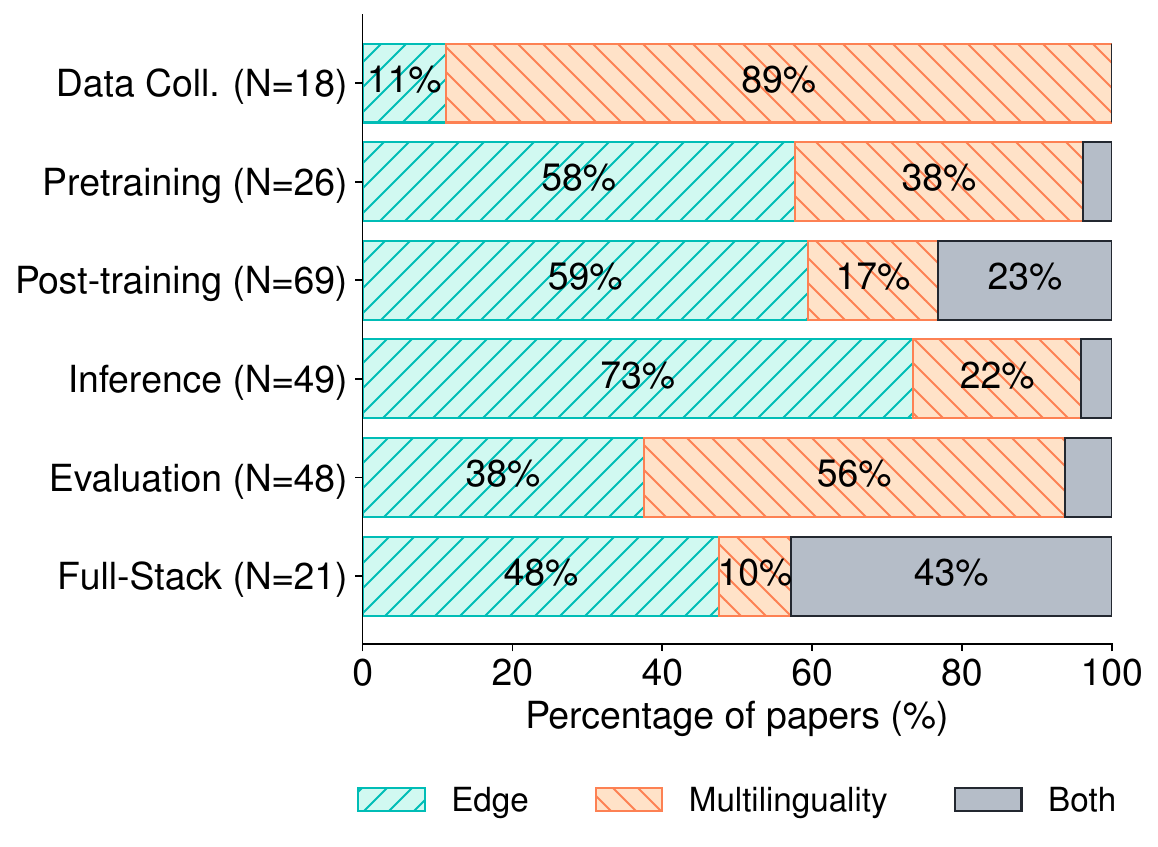}
  \caption{
    \textbf{Papers focused on Edge LM and multilinguality (N=\numpapers{}) rarely overlap within individual pipeline stages.}
    Each bar shows the share of surveyed papers classified as addressing edge LMs, multilinguality, or both.
    Full-stack work is the only category where a substantial fraction (43\%) tackles both concerns simultaneously.
    We describe our paper collection methodology in \S\ref{sec:methodology}.
  }
  \label{fig:nlp_literature_by_stage_prop}
\end{figure}

\paragraph{Opportunity: Research in these areas tends to focus on different stages of the LM pipeline}
Another reason for studying these two fields in tandem is the research \textit{opportunity} it presents.
\autoref{fig:nlp_literature_by_stage_prop} shows that individual stages of the LM pipeline tend to be dominated by one field: Data Collection and Evaluation skew toward Multilinguality, while Pretraining through Inference skew toward Edge LM development.
Looking at the papers that do focus on the full LM pipeline (Full-Stack, 9\% of total), both fields are being considered together (43\%), suggesting some evidence that end-to-end multilingual models naturally demand attention to both.
Recent works such as Tiny Aya \citep[70 languages, 3.35B,][]{salamanca2026tinyayabridgingscale}, Gemma 3n \citep[140 languages, 2B,][]{gemmateam2025gemma3technicalreport}, and Omnilingual ASR \citep[1600+ languages, with 600M param. variant,][]{omnilingualasrteam2025omnilingualasropensourcemultilingual} show that this intersection is emerging and actively being explored.


\subsection{The Language Modelling Pipeline}
\label{sec:lm_pipeline}

We ground our definition of the LM pipeline based on design choices made across recent open LM model releases such as the
OLMo \citep{walsh2025,olmo2025olmo3},
Nemotron \citep{nvidia2025nvidianemotronnano2,nvidia2025nvidianemotron3efficient}, and Aya \citep{ustun-etal-2024-aya,dang2024ayaexpansecombiningresearch,salamanca2026tinyayabridgingscale} model series.
In general, we identify five main stages: Data Collection, Pretraining, Post-training, Inference, and Evaluation.
We describe each step in \S\ref{sec:survey}.
We note that the LM pipeline is organized linearly for simplicity.
In practice, development and deployment decisions often overlap: an LM developer might
skip pretraining and start from an open-weight base model, or
evaluate iteratively during development before deploying a model for inference.
Our survey recognizes this and reflects these nuances whenever appropriate.

\subsection{Requirements of the Edge}
\label{sec:edge_constraints}

In this work, we define the \textit{Edge} as the deployment setting that imposes hardware constraints.
The edge typically involves commodity hardware such as mobile phones and laptops, and sometimes even microcontroller devices.
Edge research is relevant because much of the Global South is constrained in digital infrastructure \citep{GSMA2025MobileEconomy,occhini2026artificialintelligencecreatingnew}, and the usual routes for deploying LMs, such as using high-end GPUs or cloud services, are often inaccessible.
Successful edge deployment has been seen in domains such as healthcare \citep{hu-etal-2025-nlp-public-health-africa,rutunda-etal-2026-large}, agriculture \citep{singh2024farmerchatscalingaipoweredagricultural,samuel2025agroLLMconnectingfarmersagricultural}, and law \citep{ariai-etal-2025-nlp-legal-domain}.
Based on \citet{treviso-etal-2023-efficient}, \citet{lu-etal-2025-demystifying}, and \citet{zheng2025reviewedgeLLM}, we identify \textbf{requirements} for Edge LM deployment.

\paragraph{\iconmemory{} Memory}
The amount of memory in a device limits what LMs can be loaded and served.
It is also relevant during finetuning, where it constrains the size of a training batch.
In practice, a model's size (measured in number of parameters) and numerical precision jointly determine whether it can be loaded into memory.
For example, mobile DRAM is typically 6--12\,GB \citep{liu2024mobileLLM}, yet even a 7B-parameter model in half precision already requires over 8\,GB.

\paragraph{\iconcompute{} Compute}
The processing power of edge devices governs both the feasibility of finetuning techniques and the latency of models during inference.
Compute cost is also affected by model architecture: for example, mixture-of-experts (MoE) models tend to be more efficient than dense models because they activate only a subset of the network's parameters per input \citep{muennighoff2025olmoe}.
In addition, \citet{zheng2025reviewedgeLLM} note that the gap between edge-device capabilities and the computational demands of deploying LMs has continued to widen over time.

\paragraph{\iconenergy{} Energy}
Sustained inference can quickly drain a device's battery, restricting how long and how intensively a model can run.
For example, text generation tasks can consume 0.042 kWh, which takes 9\% of a full smartphone charge \citep{luccioni-etal-2024-power},
and even quantized LMs on a Raspberry Pi consume several joules per token \citep{husom-etal-2025-sustainable}.
A key factor in inference energy usage is the length of the generated text, and some languages are most affected because inefficient tokenization inflates their token counts \citep{ahia2023do}.

\subsection{Requirements for Multilingual Capability}
\label{sec:multilingual_constraints}

We define \textit{Multilingual Capability} as the set of methods and techniques for building language models that perform well across a broad range of languages, particularly low-resource ones.
Addressing gaps in multilinguality is important because communities in the long tail of language support also tend to face systemic constraints in digital infrastructure and socioeconomic capability \citep{occhini2026artificialintelligencecreatingnew}, further exacerbating this divide.
Based on recent surveys \citep{liu-etal-2025-culturally,longpre2025atlasadaptivetransferscaling}, we identify \textbf{requirements} to consider when building capable multilingual models.

\paragraph{\icondata{} Data}
Data has value when it's high-quality and diverse \citep{raventos2023pretraining,chen2024diversitysyntheticdataimpact}.
High-quality multilingual data enables LMs to be fluent in a community's language while maintaining strong capabilities.
Common approaches include web crawling \citep{penedo2025fineweb}, human annotation \citep{singh-etal-2024-aya}, and synthetic data generation \citep{dang2024ayaexpansecombiningresearch}.

\paragraph{\iconrepresentation{} Representation}
How text is encoded affects model performance, especially for languages with complex morphology or non-Latin scripts.
Across the LM pipeline, this is affected by design decisions such as tokenizer vocabulary \citep{rust-etal-2021-good}, input encoding \citep{minixhofer2026bolmobyteifyinggenerationlanguage}, and pretraining mixtures \citep{conneau-etal-2020-unsupervised}.

\paragraph{\iconalignment{} Alignment}
Even if a model is fluent in a certain language, it does not necessarily mean that it adheres to the cultural norms and nuances of that language's community \citep{adilazuarda-etal-2024-towards,liu-etal-2025-culturally}---cross-cultural alignment remains an unaddressed challenge.
Alignment also extends to safety, as communities differ in what they consider allowable \citep{yong-etal-2025-state}.

\begin{figure*}[t!]
  \centering
  \input{figures/taxonomy.tex}
  \caption{
    \textbf{Organization of the Survey (\S\ref{sec:survey}).}
    We organize the literature along the five stages of the LM pipeline (\S\ref{sec:lm_pipeline}),
    identifying key challenges between edge deployment constraints (\S\ref{sec:edge_constraints}, Memory \iconmemory{}, Compute \iconcompute{}, and Energy \iconenergy{})
    and properties that influence multilingual capabilities (\S\ref{sec:multilingual_constraints}, Data \icondata{}, Representation \iconrepresentation{}, and Alignment \iconalignment{}).
    Finally, we describe works that address these challenges (\S\ref{sec:survey}).
  }
  \label{fig:taxonomy}
\end{figure*}
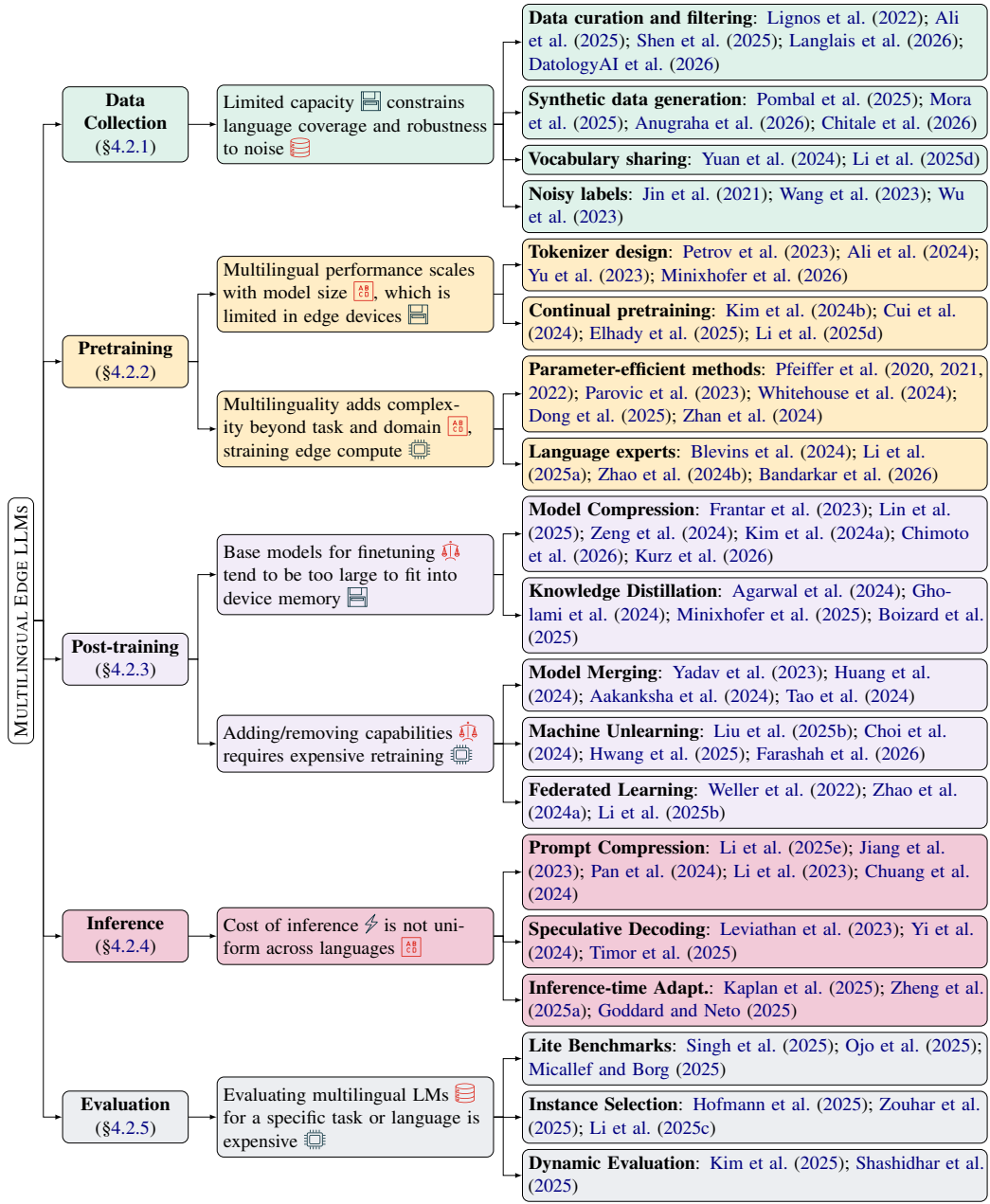

\section{Methodology}
\label{sec:methodology}

To understand the landscape of multilingual edge LMs,
we collect research from 2021 onwards on multilinguality, edge and efficient NLP, and work that combines both.
In total, we obtain \numpapers{} papers.

\paragraph{Initial Paper Screening}
First, we gather studies from peer-reviewed academic papers in $^\star$CL (e.g., ACL/EMNLP/TACL) and machine learning venues (e.g. NeurIPS/ICLR/TMLR) using the Semantic Scholar API.\footnote{\url{https://www.semanticscholar.org/product/api}}
Our search query includes keywords pertaining to multilingual NLP, efficiency, and LM deployment (\autoref{fig:search_keywords}).
This step resulted in 2,473 papers.
We screened for relevance by filtering for citations in a staggered manner, i.e., papers published in 2021 should have $\geq 100$ citations while those published in 2025 should have at least 3.

\paragraph{Filtering and Annotation}
For each work, we perform human and LM-assisted filtering and annotation to identify key attributes such as the work's
focus in the LM pipeline,
whether it focuses on multilinguality, efficiency, or both,
and deployment device.
We bootstrap annotations with GPT 4.1 Mini by including the work's title and abstract in a prompt (\autoref{fig:appendix_annotation_prompt}),
and then perform human validation to correct the annotations.

\paragraph{Final Validation}
Finally, we conduct a round of human validation to correct and refine the annotations.
In total, we collected \numpapers{} works across the whole language modelling pipeline.
We release our survey dataset in public: \href{https://huggingface.co/datasets/ljvmiranda921/multilinguality-at-the-edge}{\texttt{ljvmiranda921/\allowbreak multilinguality-at-the-edge}}

\section{Results: Multilingual Edge LM Survey}

\subsection{Overview of Surveyed Papers}

In this section, we show an overview of the papers collected for this survey.
Out of the \numpapers{} papers, 10.3\% are model releases, 76.9\% describe methodologies for developing these edge LMs, while the rest (12.80\%) are real-world system deployments.

\begin{figure}[t]
  \centering
  \includegraphics[width=\linewidth]{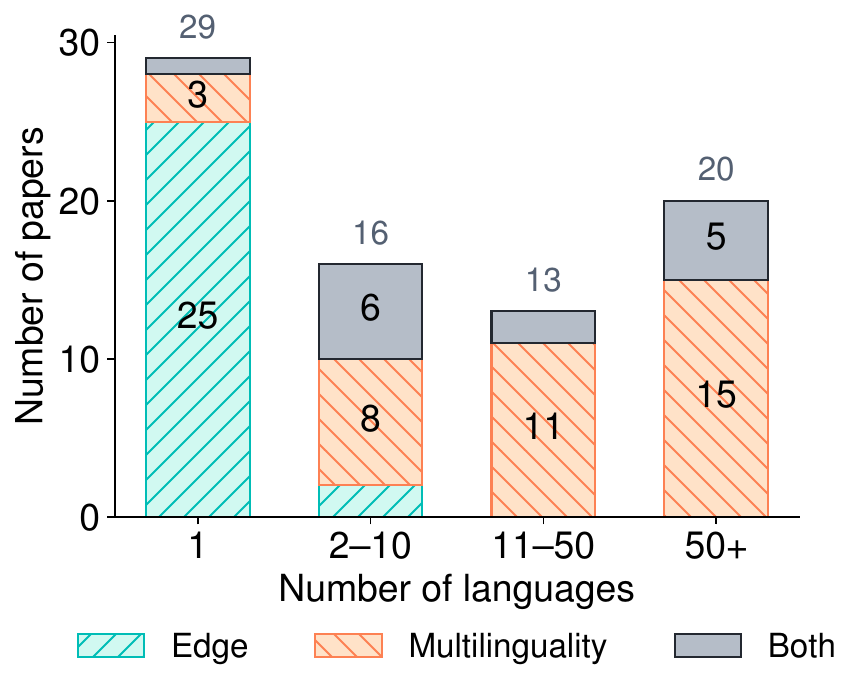}
  \caption{
    \textbf{Reported language coverage of edge LM papers.}
    We show 78 papers (of \numpapers{}) that report a concrete number of evaluated languages and bin them into four brackets:
    monolingual (1),
    few (2--10),
    many (11--50), and
    massive (50+),
    categorized by research focus.
  }
  \label{fig:language_coverage}
\end{figure}

\paragraph{Language Coverage}
Edge-only work is overwhelmingly monolingual (\autoref{fig:language_coverage}):
25 out of 29 single-language papers target edge deployment alone, while none of the edge-only papers cover more than 10 languages.
In contrast, multilinguality-focused papers skew toward broader coverage, with 15 papers supporting 50+ languages.
Papers that address both edge deployment and multilinguality remain relatively scarce, particularly at higher language counts, suggesting that the intersection of these two areas is still underexplored.

\begin{figure}[t]
  \centering
  \includegraphics[width=\linewidth, trim={2em 2em 0 0}]{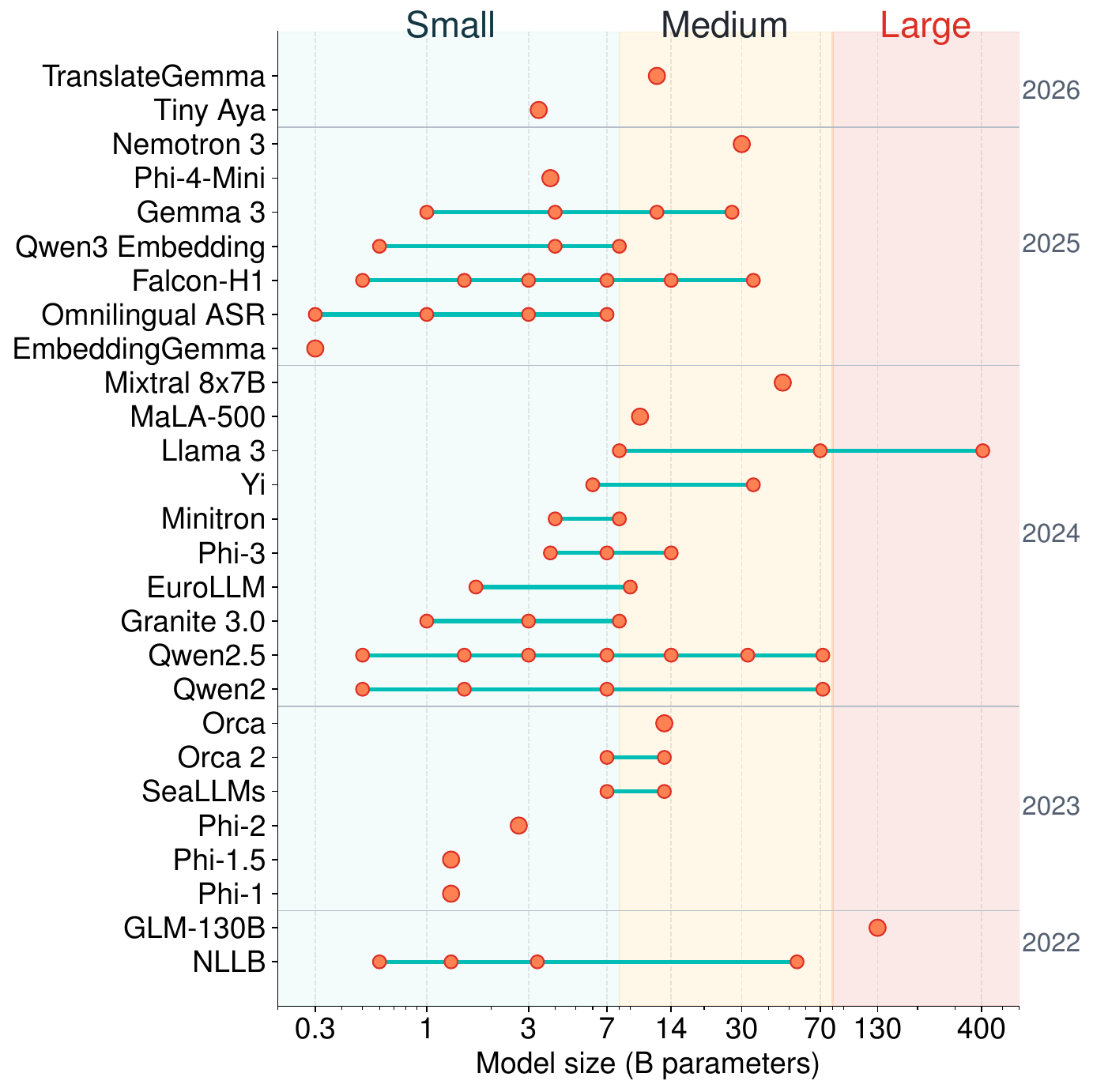}
  \caption{
    \textbf{Model sizes (in billion parameters) of various LMs.}
    For each model family in our curated set of released models, we recorded all publicly documented parameter counts and plotted the range of available sizes on a log scale.
    We adopt a simplified version of the size taxonomy proposed by \citet{jernite2026aiflopsdifferentfolks}, defining three categories:
    Small ($\leq$8B, originally `smol'), suitable for on-device deployment;
    Medium (8--80B, originally `14--32B'), targeting mid-range GPU setups; and
    Large (80B+), requiring data-center infrastructure.
  }
  \label{fig:model_sizes}
\end{figure}

\paragraph{Model Size}
We find that most model families now offer at least one variant in the small ($\leq$8B) range (\autoref{fig:model_sizes}).
Earlier releases such as GLM-130B and NLLB (2022) were concentrated in the medium-to-large regime,
but from 2023 onward, model families increasingly span a wider range of sizes, with small variants becoming the norm.
For example, Qwen2.5 and Gemma 3 both offer variants from under 1B to over 30B parameters.
This suggests a growing emphasis on making multilingual models accessible for edge deployment.

\subsection{Challenges and Methods for Building Multilingual Edge Language Models}
\label{sec:survey}

We now examine \textbf{challenges}, i.e., how the constraints of edge deployment and the requirements of multilinguality compete at each stage of the LM pipeline,
and survey different \textbf{methods} that address them.
An overview is shown in \autoref{fig:taxonomy}.

\subsubsection{Data Collection}
\label{sec:data_collection}

The LM pipeline often begins with sourcing and curating text corpora for both pretraining and post-training.
Pretraining corpora tend to be unstructured web-crawled text, while post-training datasets are more structured.

\paragraph{Challenge: Limited model capacity \iconmemory{} limits language coverage while being susceptible to noise \icondata{}}
The parameter-sharing paradigm prevalent in LMs imposes a limit on their capacity, creating two challenges for multilingual edge models.
First, adding more languages dilutes the model's capacity for existing ones \citep{doddapaneni2025primer},
a phenomenon often dubbed the curse of multilinguality \citep[\textit{inter alia}]{chang-etal-2024-multilinguality,longpre2025atlasadaptivetransferscaling}.
Regional multilingual models may cover 10--20 languages \citep{pava2025mind}, such as SEA-LION \citep{ng2025sealionsoutheastasianlanguages} for Southeast Asia or Updesh \citep{chitale2026updeshsynthesizinggroundedinstruction} for Indic languages, yet scaling further risks degrading per-language quality.
Second, this limited capacity renders a model more sensitive to noise \citep{havrilla2024understandingeffectnoiseLLM}, and low-resource languages suffer from this exact problem \citep{kreutzer-etal-2022-quality},
as noise manifests in both pretraining \citep{chen2024understanding,longpre-etal-2024-pretrainers} and post-training \citep{ijcai2024p403,zhu-etal-2024-fine} data.
Addressing this challenge requires, given a fixed parameter budget, maximizing language coverage without sacrificing data quality.

\paragraph{Methods}
Getting the right mixture of languages in the training dataset is effective for mitigating this challenge.
Several approaches leverage \textbf{vocabulary sharing}, i.e.,  selecting languages for the data mix based on whether their inclusion boosts the performance of other languages \citep[\textit{inter alia}]{yuan-etal-2024-vocabulary,li2025rethinking}.
\textbf{Synthetic data generation} can also induce an optimal mixture:
by choosing a strong teacher model \citep{miranda2026polyglotteachersevaluatinglanguage},
crafting prompts \citep{mora2025artaskingmultilingualprompt},
and performing quality validation \citep{anugraha2026mr3multilingualrubricagnosticreward,pombal2025mprometheus},
one can produce high-quality training data for small multilingual models \citep{devine2026kakugodistillationlowresourcelanguages,kim2026dataefficientpretrainingscalingsynthetic}.
\textbf{Data curation and filtering} remains a reliable and high-leverage intervention that involves careful filtering, deduplication, and quality control of multilingual web corpora \citep{kuduguntua-etal-2023-madlad400,langlais2026common,datologyai2026uberwebinsightsmultilingualcuration},
adapted to multilingual contexts \citep{ali-etal-2025-judging,shen2025dcad} and common data sources such as Wikipedia \citep{lignos-etal-2022-toward}.
When working with low-resource annotation pipelines, handling \textbf{noisy labels} through noise-robust training methods is also effective \citep{jin-etal-2021-instance,wang-etal-2023-noise,wu-etal-2023-noisywikihow}.
These methods are largely data-centric; in \S\ref{sec:pretraining}, we discuss how this challenge can also be addressed by expanding the model's capacity.

\subsubsection{Pretraining}
\label{sec:pretraining}

Web-crawled text is then used to train a base model to learn general language representations.
This is often the most computationally expensive stage;
for example, \citet{nvidia2025nvidianemotronnano2} pretrain on 20T tokens, while \citet{olmo2025olmo3} use 8$\times$ NVIDIA H100s for $\sim$6T tokens.

\paragraph{Challenge 1: Multilingual performance scales with model size \iconrepresentation{}, which is limited in edge devices \iconmemory{}}
Hardware imposes a ceiling on the size of the model that can be deployed.
\citet{jernite2026aiflopsdifferentfolks} note that small LMs ($\leq$8B parameters) can run on mobile phones, low-end GPUs, and even CPUs for smaller workloads, though they still significantly increase the compute load of local devices and require continued hardware progress.
Moreover, multilingual scaling laws show that performance improves with model size \citep{longpre2025atlasadaptivetransferscaling}, creating a tension between achieving strong multilingual performance and fitting within the constraints of a local device.
Addressing this challenge requires careful tradeoffs: for instance, some approaches forego language coverage or capability in favor of language- or task-specific edge LMs.

\paragraph{Methods}
Model parameter sizes (e.g., 7B, 32B, 70B) are largely set by design choices prior to pretraining \citep{kaplan2020scalinglawsneurallanguage}, but there are ways to ensure that a fixed parameter budget still supports strong multilingual performance.
One is through \textbf{tokenizer design}: choices such as subword allocation \citep{petrov2023language,ali-etal-2024-tokenizer} and byte-level encoding \citep{yu-etal-2023-megabyte,minixhofer2026bolmobyteifyinggenerationlanguage} affect how a fixed model represents diverse languages.
Another is \textbf{continual pretraining}, which involves vocabulary expansion \citep{kim2024efficienteffectivevocabularyexpansion,cui2024efficienteffectivetextencoding} and further training on target-language data to adapt an existing model to new languages \citep{elhady-etal-2025-emergent,li2025rethinking} without retraining from scratch.
The latter has been a common approach in most regional LMs such as in Southeast Asian \citep{ng2025sealionsoutheastasianlanguages} and Japanese \citep{fujii2024continual} languages.

\paragraph{Challenge 2: Multilinguality adds complexity beyond task and domain \iconrepresentation{}, straining edge compute \iconcompute{}}
Language models must already capture variation across tasks and domains.
Multilinguality introduces yet another axis of complexity such as morphology and orthography, to name a few \citep{tsvetkov2017lowresource}.
In a standard dense architecture, all parameters are activated for every input regardless of language, making compute costs proportional to total model complexity.
This is particularly problematic on edge devices, where mobile GPUs peak at 2--6.5 TFLOPS (FP16) with 26--63 GB/s memory bandwidth \citep{xiao2026llmpockets}, orders of magnitude below datacenter capacity.

\paragraph{Methods}
Minimizing an LM's compute footprint while retaining performance often involves architectural modifications or training strategies beyond the standard dense model.
\textbf{Parameter-efficient methods} have been applied in multilingual settings across a variety of languages and tasks.
These involve techniques such as adapters \citep{houlsby-etal-2019-parameter,pfeiffer-etal-2020-mad,pfeiffer-etal-2021-unks,pfeiffer-etal-2022-lifting,parovic-etal-2023-cross}, LoRA \citep{hu2022lora,whitehouse-etal-2024-low,dong-etal-2025-mlas,owodunni2025continuallyaddingnewlanguages}, and prefix tuning \citep{li-liang-2021-prefix,zhan-etal-2024-prefix}.
Architectural interventions such as \textbf{language experts} have also been explored, where strong monolingual LMs are trained and combined through gating \citep{blevins-etal-2024-breaking,li-etal-2025-group} or routing \citep{zhao2024sparse,bandarkar2026multilingualroutingmixtureofexperts}.
Compared to a dense model, this class of methods only activates a subset of parameters per input while preserving capacity.

\subsubsection{Post-training}
\label{sec:post_training}

A base model is further adapted toward specific capabilities through techniques such as supervised finetuning \citep[SFT,][]{ouyang2022traininglanguagemodelsfollow} or reinforcement learning.
We also group the literature on model compression and merging under this stage.
This stage is typically less computationally expensive than pretraining, and often most language communities start development at this stage by finetuning an existing base model.

\paragraph{Challenge 1: Models for finetuning \iconalignment{} tend to be too large to fit into memory \iconmemory{}}
Since pretraining is often expensive, practitioners resort to finetuning existing open-weight models released on platforms such as HuggingFace \citep{davidson2023aicapabilitiessignificantlyimproved,wolfe-etal-2024-laboratory}.
However, adapting a pretrained base LM means inheriting its architectural decisions, including model size.
If the base LM exceeds the memory constraints of the target edge device \citep{zheng2025reviewedgeLLM}, the finetuned model will as well.
Addressing this may involve compressing the post-trained model, or distilling the large model's knowledge into a smaller one.

\paragraph{Methods}
Compressing a model can happen before, during, or right after the post-training stage.
\textbf{Model compression} attempts to reduce the size of a model while retaining its capabilities.
This can be achieved through quantization \citep{frantar2023optq,lin-etal-2025-awq}, which expresses model weights in lower precision, or
pruning, which removes redundant weights from the network \citep{zeng-etal-2024-multilingual,kim-etal-2024-pruning}.
Both methods rely on a calibration dataset to guide the compression process, and several works find that tailoring this dataset to the target language is important for preserving multilingual performance \citep{chimoto-etal-2026-calibrating,kurz-etal-2026-limitations}.
\textbf{Knowledge distillation} is another method that transfers the capabilities of a larger LM (teacher) into a smaller LM (student) by training the student to mimic the teacher's behavior.
Generating synthetic data from the teacher is one approach as discussed in \S\ref{sec:pretraining} (off-policy), but it is also possible to learn this on-the-fly during training---an approach called on-policy distillation \citep[OPD,][]{agarwal2024onpolicydistillationlanguagemodels}.
However, OPD typically requires the teacher and student to share the same tokenizer, and several works have attempted to address this restriction \citep{gholami-etal-2024-gold,minixhofer2025universal,boizard2025towards}.

\paragraph{Challenge 2: Adding or removing capabilities \iconalignment{} requires expensive retraining \iconcompute{}}
If the base model lacks desired capabilities or retains extraneous knowledge from pretraining,
the naive approach is to retrain from scratch, which can be expensive on resource-constrained hardware.
Adding capabilities, such as support for a new language, is the more obvious case.
However, removing capabilities is also important: for example, on-device models may need to unlearn sensitive or private information from pretraining to comply with data regulations or to free up capacity \citep{habernal-etal-2023-privacy}.
Addressing this challenge requires performing such activities while reducing (or fully-eliminating) training costs.

\paragraph{Methods}
Adding capabilities with minimal retraining can be achieved via \textbf{model merging} \citep{yadav-etal-2023-tiesmerging}, which has shown to be effective in multilingual scenarios \citep[\textit{inter alia}]{huang-etal-2024-chat,aakanksha2024mix,tao-etal-2024-unlocking}.
For removing capabilities, \textbf{machine unlearning} \citep{liu-etal-2025-rethinking} enables a model to forget specific knowledge without full retraining, for example by fine-tuning on synthetic data that overwrites sensitive information \citep{yu2022differentially,yue-etal-2023-synthetic,yu-etal-2024-privacy} or by performing model editing at a parameter level \citep{choi-etal-2024-cross,hwang2025uncoveringpotentialrisksunlearning,farashah2026multilingualamnesiatransferabilityunlearning}.
\textbf{Federated learning} addresses both directions: it allows distributed clients to collaboratively add language capabilities to a shared model while keeping private data on-device.
While initial work explored this in the pretraining setting \citep{weller-etal-2022-pretrained}, federated approaches are particularly suited to post-training, where community-held data can be used to finetune or align a base model without requiring centralized collection \citep{zhao2024breaking,li-etal-2025-multilingual-federated}.

\subsubsection{Inference}
\label{sec:inference}

At this stage, a model is served on a target device, which can range from microcontrollers to data center-scale GPUs in terms of device capabilities.
Inference can occur either online, where a model runs on a remote server and is accessed via an API, or offline, where the model runs directly on the user's device.
For the Global South, online inference is constrained by network connectivity \citep{itu-unesco-2025-state}, while offline inference is limited by memory (whether the model fits on the device) and energy (whether the device can sustain extended workloads).

\paragraph{Challenge: Cost of inference \iconenergy{} is not uniform across languages \iconrepresentation{}}
Several works have shown that inference accounts for a significant share of an LM's total energy consumption across its lifecycle \citep[\textit{inter alia}]{fu-etal-2025-llmco2,adamska2025greenprompting}, with some estimates reaching 90\% \citep{hutt2019inferentia}.
This has a considerable effect on device sustainability and overall emissions.
The length of a model's input or output is directly proportional to this cost \citep{jegham2025hungryaibenchmarkingenergy}, which in multilingual contexts is affected by how efficiently a model tokenizes a language \citep{rust-etal-2021-good,ahia2023do,petrov2023language}.
Since a model might tokenize languages differently, a ``token tax'' \citep{lundin-etal-2026-token} is often accrued per query.
While interventions during development can resolve these issues (\S\ref{sec:pretraining}--\ref{sec:post_training}), addressing this challenge also requires inference-time solutions.


\paragraph{Methods}
At inference, practitioners can reduce costs through two levers: making prompts more efficient or adjusting the deployed model's tokenization scheme.
\textbf{Prompt compression} addresses prompt efficiency by removing unnecessary tokens from the original prompt while preserving its meaning \citep{li-etal-2025-prompt}.
A notable line of work is the LLMLingua series \citep{jiang-etal-2023-llmlingua,pan-etal-2024-llmlingua,jiang-etal-2024-longllmlingua}, among others \citep{li-etal-2023-compressing,chuang-etal-2024-learning}.
\textbf{Speculative decoding} is another class of methods that uses a smaller LM (drafter) to predict potential future tokens \citep{leviathan-etal-2023-fast}.
\citet{yi-etal-2024-towards} showed that tuning these drafters to the target language can improve both performance and inference time in multilingual tasks such as translation.
Although speculative decoding typically requires that the drafter and deployed LM share the same vocabulary, \citet{timor2025accelerating} showed that this requirement can be lifted.
Finally, \textbf{inference-time adaptation} addresses the second lever by improving the deployed LM's tokenization efficiency or vocabulary for the target language without retraining.
For example, \citet{kaplan2025from} and \citet{zheng2025broken} exploit an LM's internal representation to enable vocabulary expansion or handle unseen tokenization.
Some works approach this problem by approximating out-of-vocabulary tokens and transferring this knowledge without retraining \citep[\textit{inter alia}]{goddard2025trainingfreetokenizertransplantationorthogonal}.

\subsubsection{Evaluation}
\label{sec:evaluation}

Model performance is measured across capabilities and domains, either on static benchmarks or on structured testbeds. 
Evaluation can also happen during development to guide decisions before deployment.
For multilingual evaluation, benchmarks are typically either task-specific, evaluating a single capability across many languages \citep[\textit{inter alia}]{shi2023language,singh-etal-2025-global,gureja-etal-2025-rewardbench}, or
language-specific, evaluating a language (or a group of closely-related languages) across many tasks \citep{ojo-etal-2025-afrobench,miranda-etal-2025-filbench}.

\paragraph{Challenge: Evaluating multilingual LMs \icondata{} for a specific task or language is expensive \iconcompute{}}
Choosing or updating a model for deployment requires concrete quantitative evidence.
However, these evaluations can significantly increase compute costs, as benchmarks often contain thousands of instances spanning many languages and capabilities.
For example, Global-MMLU \citep{singh-etal-2025-global} contains 14.3k instances across 42 languages, while AfroBench \citep{ojo-etal-2025-afrobench} and FilBench \citep{miranda-etal-2025-filbench} span $>100\text{k}$ instances.
Although most of these works provide recommendations on which LMs perform best at the time, they can easily get outdated as new LMs are released \citep{reiter2026comparing}.
We consider this challenge an essential part of deployment because evaluating a model is quite prominent in the edge \citep{ramjee-etal-2025-ashabot,rutunda-etal-2026-large}, especially for context-specific use-cases.
Addressing this challenge requires not only creating smaller and high-quality benchmarks, but also designing holistic approaches to evaluation outside of static datasets.

\paragraph{Methods}
In order to reduce the number of instances for evaluation,
researchers have developed \textbf{lite benchmarks} as companions to the standard benchmarks they released \citep{singh-etal-2025-global,ojo-etal-2025-afrobench,micallef-borg-2025-melabenchv1}.
These versions are intended to speed up evaluation during iterative development, rather than to serve as the final leaderboard scores reported at model release.
Other techniques borrow from established methods such as item response theory (IRT) and reinforcement learning (RL) to perform \textbf{instance selection}, identifying the most informative items for evaluation \citep{hofmann2025fluid,zouhar-etal-2025-how,li2025active}.
Finally, another class of techniques create \textbf{dynamic evaluation} either from an existing evaluation dataset \citep{kim2025benchhub} or a set of unstructured documents \citep{shashidhar2025yourbench}.

\section{Analysis: From Methods to Systems}

Thus far, we have discussed the challenges and potential methods that researchers and practitioners may encounter when developing and deploying multilingual edge LMs.
In this section, we turn to completed edge LM efforts that have been integrated into real-world applications, which we refer to as \textbf{edge LM systems}, and examine
how they are made (\S\ref{sec:how_are_edge_lm_systems_made}),
who develops them (\S\ref{sec:who_develops_edge_lm_systems}),
and which domains they are deployed to (\S\ref{sec:which_domains_are_edge_lm_systems}).
An example of this is \citet{ramjee-etal-2025-ashabot}, describing how an industry laboratory, NGO, and academic institution collaborated to deploy an LM-based chatbot (ASHABot) for community health workers in Rajasthan, India.
In order to identify an edge LM system,
we manually classify each of the \numpapers{} papers on whether an actual model deployment took place (yes/no), obtaining \numdeploymentpapers{} papers in the process.

\subsection{How Are Edge LM Systems Made?}
\label{sec:how_are_edge_lm_systems_made}

\begin{figure}
  \centering
  \includegraphics[width=\linewidth, trim={0 0.5cm 0 0}]{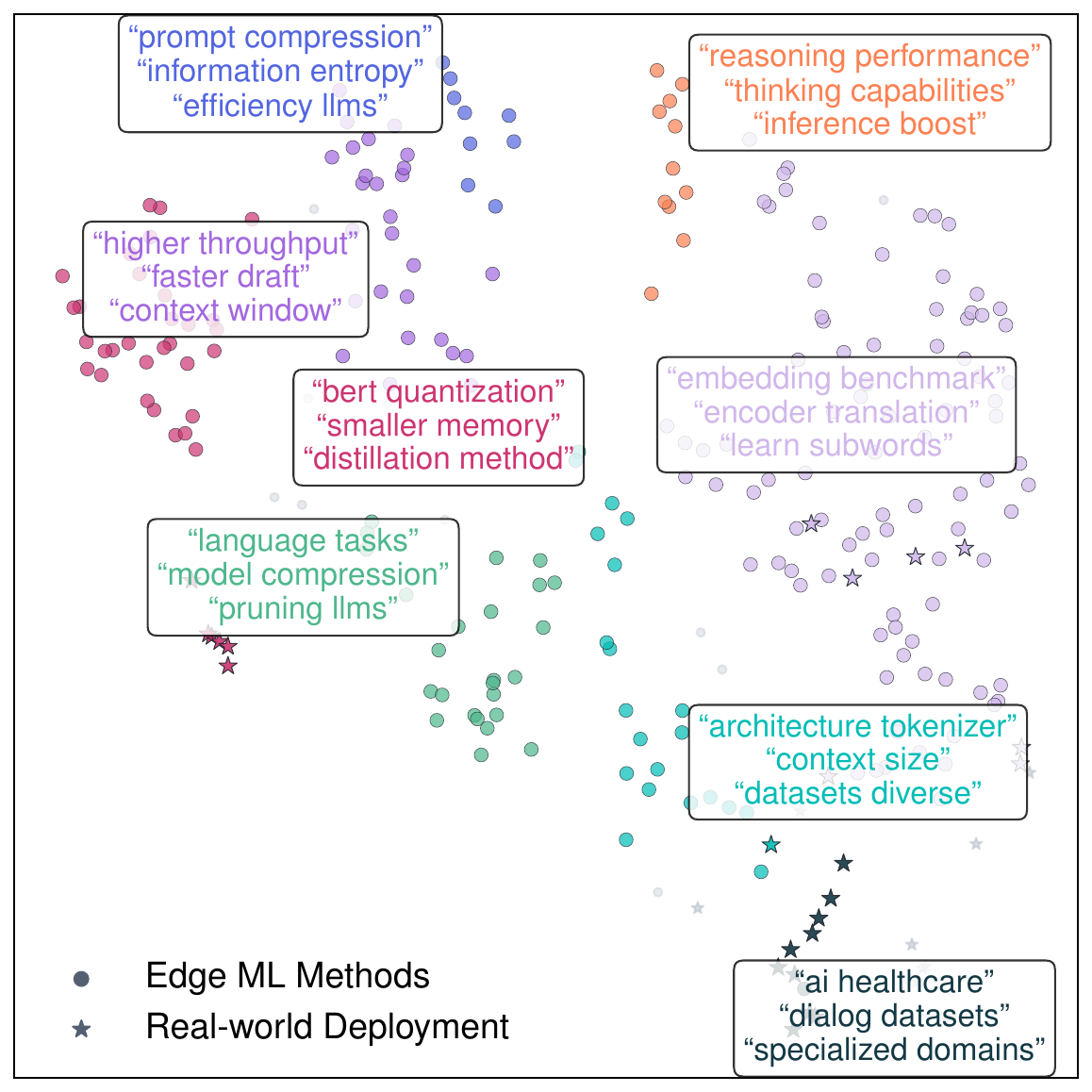}
  \caption{
    \textbf{Clustering of \numpapers{} surveyed papers by abstract similarity.}
    Although real-world deployments ($\star$, n=\numdeploymentpapers{}) are present in some clusters, they tend to concentrate near select keywords
    (e.g., dialog datasets, distillation method, quantization, etc.),
    suggesting that real-world deployments share common methodological characteristics.
  }
  \label{fig:literature_clusters_map}
\end{figure}

\paragraph{Setup}
In order to examine how edge LM systems are made, we manually tag the methods each system uses across the LM pipeline (we show a representative sample in \autoref{tab:lm_pipeline_edge_lm_systems}).
We also situate these works within the broader \numpapers{} surveyed papers by visualizing their embeddings.
We achieve this by embedding all \numpapers{} abstracts using MiniLM \citep{wang-etal-2020-minilm},
clustering them using UMAP and HDBSCAN \citep{mcinnes-etal-2017-hdbscan,micinnes-etal-2018-umap},
and extracting keywords for each cluster using KeyBERT \citep{grootendorst2020keybert}.

\paragraph{Findings}
\autoref{fig:literature_clusters_map} shows the embeddings (reduced to 2-dim) of each paper, with edge LM systems marked.
These deployments concentrate near a few clusters such as ``model compression'' and ``dialog datasets,'' while clusters like ``reasoning performance'' or ``prompt compression'' have little to no representation.
We also find that synthetic data generation is a prevalent approach for data collection, while model compression via quantization is common during post-training.
This suggests that edge LM deployments favor a relatively narrow set of methods, leaving significant opportunities to explore alternatives.

\begin{figure*}[t]
  \centering
  \begin{subfigure}[t]{0.48\textwidth}
    \centering
    \includegraphics[width=\linewidth,trim={0 0 0 0}]{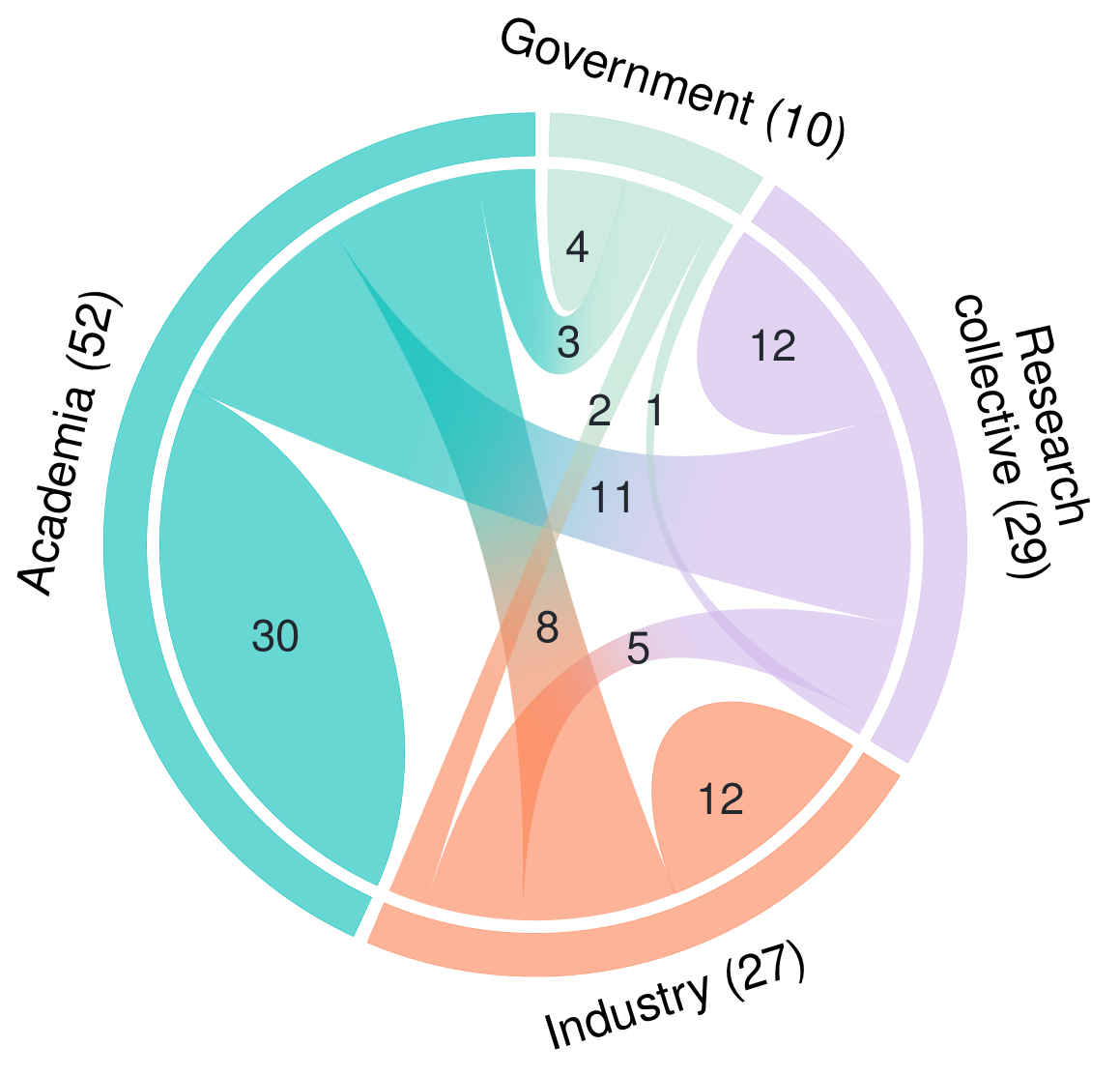}
    \caption{
      \textbf{Affiliation type of authors from papers that deployed edge LM systems.}
      Numbers on the chords show how many papers are shared between (or within) sectors.
      Academia has the largest proportion of collaborations, while government participation remains limited.
    }
    \label{fig:collaboration_sectors}
  \end{subfigure}
  \hfill
  \begin{subfigure}[t]{0.48\textwidth}
    \centering
    \includegraphics[width=0.90\linewidth, trim={0 1.7cm 0 0}]{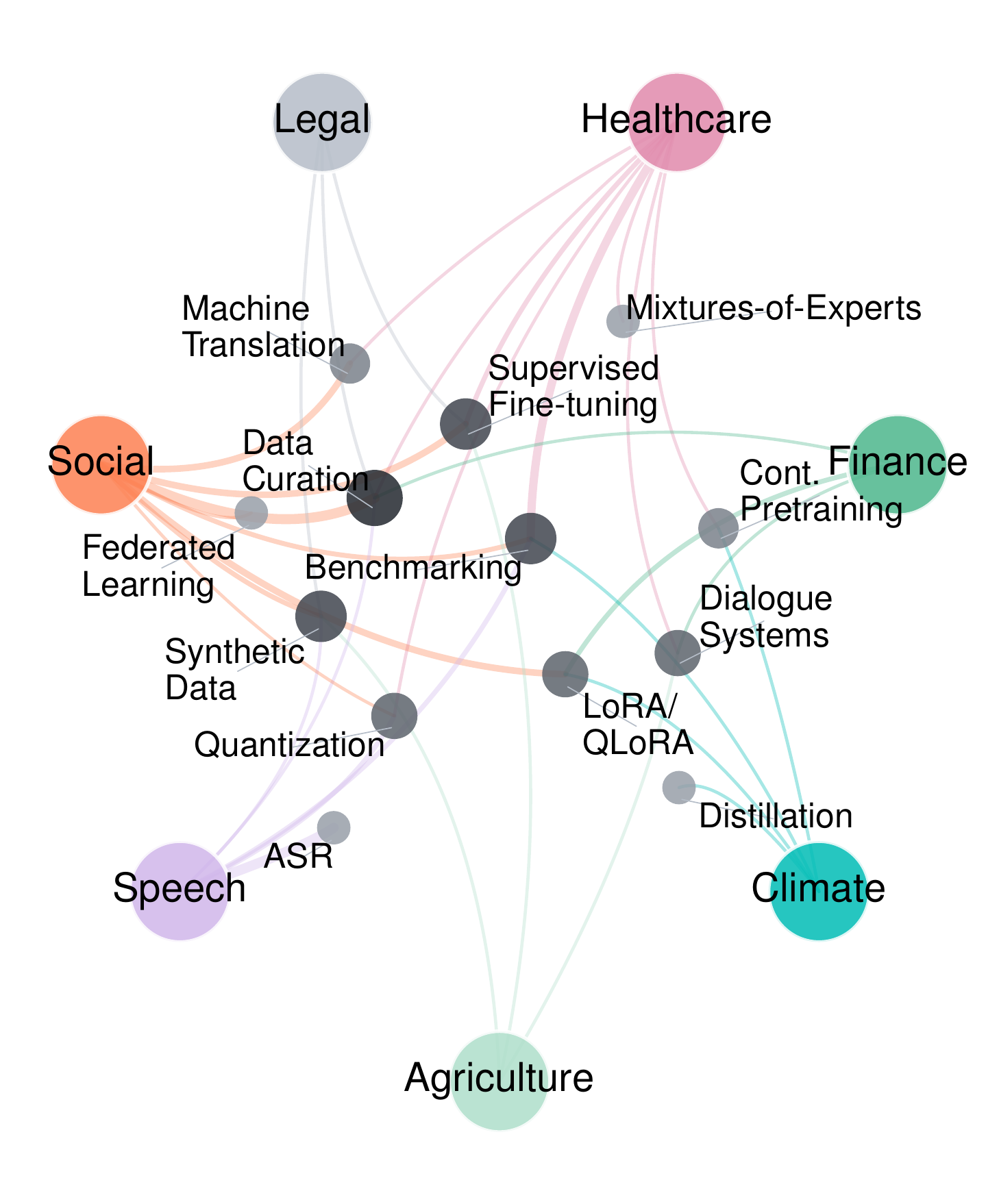}
    \caption{
      \textbf{Edge LM real-world deployment domains network.}
      Central nodes represent methods used to develop and deploy real-world edge LMs.
      Edge color indicates connectivity, while darker nodes indicate high sharing among domains.
    }
    \label{fig:domain_method_network}
  \end{subfigure}
\end{figure*}

\subsection{Who Develops Edge LM Systems?}
\label{sec:who_develops_edge_lm_systems}

\paragraph{Setup}
In order to identify who develops edge LM systems, we manually classify the affiliations of the authors from the \numdeploymentpapers{} system papers
according to various \textbf{sectors} in \citet{maslej2025artificialintelligenceindexreport}'s taxonomy, which includes
Academia (universities and affiliated research institutions),
Government (state-affiliated institutes or public sectors like hospitals),
Industry (ranges from startups to enterprise companies that release frontier LMs),
and Research collective (non-profit research organizations).
For authors with multiple affiliations, each affiliation is counted separately.
We then measure cross-sector \textit{collaborations} by counting how often each pair of sectors co-occurs within the same paper.

\paragraph{Findings}
\autoref{fig:collaboration_sectors} shows a chord diagram of different sectors and their collaborations across the \numdeploymentpapers{} papers.
Academia is well-represented, followed by research collectives and the industry.
Cross-sector collaborations are also common especially across academia and research collectives.
Government participation remains limited and is mostly driven by cross-sector collaborations with academia.
This suggests that most edge LM deployments involve cross-sector collaborations, likely because the combined challenges of hardware and capabilities benefit from complementary expertise.
It also indicates that they are being developed for real-life applications that benefit from such collaborations.

\subsection{Which Domains are Edge LM Systems Deployed to?}
\label{sec:which_domains_are_edge_lm_systems}

\paragraph{Setup}
In order to map the domains in which an edge LM is deployed,
we perform a round of classification by tagging each paper according to their \textbf{domain}, based on the following categories from \citet{rogers-etal-2023-qa} and \citet{chen2024a}: Agriculture, Climate, Finance, Healthcare, Legal, Social, and Speech.
Then, we extract mentions of different methods by keyword matching via KeyBERT, and visualize the domain-method connections as a network graph.

\paragraph{Findings}
\autoref{fig:domain_method_network} shows a network visualization linking multilingual edge LM deployment domains to the methods mentioned in their abstracts.
We find that some methods are consistently used across domains, such as synthetic data generation, data curation, SFT, and LoRA.
Healthcare and Social exhibit the broadest coverage (most edges), while Legal and Agriculture connect through fewer, more generic methods.
This suggests that the diversity of methods in edge LM systems is uneven across domains, with some seeing broad experimentation while others remain concentrated around a few common techniques.

\section{Discussion: Reaching the Last Mile}
\label{sec:discussion}

We discuss what it takes (open challenges and recommendations) for the field to reach the \textit{last mile} and deploy useful LMs
nearest to the communities that need these technologies the most.

\subsection{Development in the edge}
Throughout this work, we assume that models are developed in a resource-rich environment such as a university research cluster or industry laboratory, and then deployed to the edge.
But development can also happen on the edge itself, given that communities on the edge have the knowledge and unique perspective of their contexts
and might prefer to be involved as active contributors in research teams \citep{pillai-etal-2023-community}.
In addition, communities might value sovereignty, or their ownership of their data and models \citep{jiang2024models}.

Model development requires readiness, and \citet{occhini2026artificialintelligencecreatingnew} define a community's AI readiness through the confluence of three factors: data and model resources, digital infrastructure, and socioeconomic capability.
Our work maps to the first two---multilingual capabilities (\S\ref{sec:multilingual_constraints}) as a function of data resources, and edge constraints (\S\ref{sec:edge_constraints}) as digital infrastructure.
Although socioeconomic constraints remain unaddressed in this work,
we find that communities bypass these through research collectives, which we recommend as an investigation left for future work.

\subsection{Open Challenges and Future Work}

In this section, we discuss open challenges that go beyond the pipeline-level concerns in our survey (\S\ref{sec:survey}) and outline directions for future work.

\paragraph{Extremely Low-Resource (XLR) Languages}
Most of the methods in the data collection section (\S\ref{sec:data_collection}) work well for languages that have decent representation on the Internet.
However, the extremely low-resource (XLR) regime is characterized by $\leq$10M tokens \citep{conneau-etal-2020-unsupervised} or languages with no established orthography or written tradition \citep{pava2026digitization}
that are not easily mitigated by naively curating or synthesizing data.
Some works address this by tuning the synthetic data pipeline for the target language, e.g., by optimizing the prompt \citep{mora2025artaskingmultilingualprompt} or the underlying generator model \citep{mitra2024agentinstructgenerativeteachingagentic}.
Ultimately, the best way to drive innovations in data collection for these languages is to include the communities that speak them, going beyond simply crawling the Web.
For example, we see efforts to build collaborative human-AI tools to obtain grounded datasets for LM training \citep{f-p-dossou-aidasso-2025-towards} or application of methods such as reinforcement learning to learn from limited data \citep{sutawika2026gainedtranslationprivilegedpairwise,attia2026improvinglowresourcemachinetranslation}.

\paragraph{Orchestrating Edge LMs}
Throughout this work, we assume that a multilingual edge LM is a single model capable of performing a downstream task (e.g., chat assistant, translation, etc.).
However, agentic paradigms explore how to improve performance by orchestrating a variety of tools and models.
While coding and mathematical reasoning have benefited from these approaches, we posit that the challenges in the Global South are also well-suited for them.
For example, an orchestrator model could integrate strong multilingual LMs, expert-created tools, and translation models to solve a complex task \citep{nielsen2026learning}.
Since these orchestrator models only need to perform a single verifiable task (i.e., tool-calling), they can be made smaller \citep{erdogan-etal-2024-tinyagent,belcak2025smalllanguagemodelsfuture}.
The challenge then lies in ensuring that these orchestrator LMs invoke the right tools or LMs, figuring out the right mix of tools to put in the orchestrator's context, and finetuning them to understand the problem at hand.

\paragraph{Conceptual Frameworks for \textit{Tackling} Overlapping Constraints}
There are many conceptual frameworks in NLP that describe this problem of overlapping constraints between data and infrastructure for the Global South: the low-resource double bind \citep{ahia-etal-2021-low-resource}, square-one bias \citep{ruder-etal-2022-square}, and Zeno's paradox \citep{nigatu-etal-2024-zenos}, among others.
However, there are fewer conceptual frameworks that seek to address these problems.
We surmise that such frameworks can be drawn from other fields, such as economics or natural sciences, where reasoning about overlapping constraints is the norm.
For example, \citet{hausmann2005growth} from development economics posit that despite overlapping constraints, one only needs to identify and address the most binding constraint.
Future work may involve adapting these frameworks to the context of LM development and examining how they map to real-world use-cases.




\subsection{Recommendations to Stakeholders}

Cognizant of the fact that multilingual capabilities and edge constraints interact at various stages of the LM pipeline,
we now outline our recommendations to different stakeholders in order to develop and deploy capable edge LMs.

\begin{itemize}[leftmargin=5mm,itemsep=0mm]
  \item \textbf{For NLP researchers and model developers:}
        First, it is important that \textit{evaluation of edge models considers other constraints} (\S\ref{sec:edge_constraints}) such as energy or compute.
        Our analysis in \autoref{tab:lm_pipeline_edge_lm_systems} shows that most edge LMs focus on memory (parameter size), yet several communities might face other constraints that prohibit access to these tiny models.
        Also, our finding that edge deployments cluster around a narrow set of methods (\autoref{fig:literature_clusters_map}) suggests that
        \textit{exploring underrepresented methods} could yield better edge LMs.
  \item \textbf{For deployment practitioners and communities at the edge:}
        We find that successful edge LM deployments often involve research collectives partnering with other sectors (\autoref{fig:collaboration_sectors}).
        Thus, \textit{encouraging these types of partnerships} is paramount.
        For example, ASHABot \citep{ramjee-etal-2025-ashabot} involved collaboration between an industry laboratory, NGO, and academic institution to deploy a healthcare chatbot in India, while \citet{rutunda-etal-2026-large} followed a similar model in Rwanda.
        At the same time, we echo \citet{pillai-etal-2023-community} and \citet{petti2025coact} in saying that \textit{communities on the edge should take a more active role} in development,
        not as consultants, but as collaborators.
  \item \textbf{For policymakers and funders:}
        \autoref{fig:infra_lingdiv_ict} shows that 12 out of 21 countries with 100+ living languages are low or middle-income, suggesting that funding should point not only towards model development but also towards the infrastructure and devices that make these deployments accessible.
        In addition, we observe from our surveyed papers that government participation is limited (\autoref{fig:collaboration_sectors}), reinforcing the need for more involvement.
\end{itemize}

Finally, we highlight that reaching the \textit{last mile} does not mean that it is a problem that can be solved once and for all.
We show in our analyses and recommendations that this process is continuous, requiring consistent interactions among stakeholders across domains.
To that end, we encourage stakeholders to revisit and build upon our findings as the landscape of edge deployment evolves.



\section{Conclusion}

We examined the challenge of the \textit{last mile}: deploying LMs nearest to the communities that need them the most.
This means building models that serve languages beyond English while remaining small and fast enough for hardware-constrained environments.
Bridging these two requirements introduces challenges at every stage of the LM pipeline, and we surveyed \numpapers{} papers that propose methods to address them.
We hope that this work paves the way for more equitable language technologies.

\iftaclpubformat
\section*{Acknowledgments}
This work is supported by the UK Research and Innovation (UKRI) Frontier Research Grant EP/Y031350/1 (EQUATE) awarded to AK at the University of Cambridge.
LJVM would also like to thank the Microsoft Research Grant for the compute credits used to access GPT-4.1.
\else
\fi

{
  \small
  \bibliography{tacl2021}
  \bibliographystyle{conf_acl_natbib}
}

\clearpage
\appendix


\begin{figure*}[ht]
  \small
  \centering
  \begin{minipage}{0.95\textwidth}
    \promptbox[\textbf{Search Keywords:} Keywords for the Initial Paper Screening Step]{
      \texttt{
        ("multilingual" | "cross-lingual" | "crosslingual" | "low-resource language*" | "polyglot" | "code-switching" | "language
        transfer" | "zero-shot cross-lingual" | "massively multilingual" | "underrepresented language*" | "endangered language*" |
        "minority language*" | "language-agnostic" | "compression" | "quantization" | "pruning" | "distillation" | "edge" |
        "on-device" | "tinyml" | "lightweight" | "small language model*" | "knowledge distillation" | "model compression" | "weight
        sharing" | "neural architecture search" | "efficient inference" | "mobile NLP" | "sparse model*" | "mixture of experts" |
        "low-rank" | "LoRA" | "parameter-efficient" | "PEFT" | "vocabulary pruning" | "speculative decoding" | "early exit" |
        "multilingual distillation" | "cross-lingual transfer" | "vocabulary reduction" | "tokenizer" | "subword" | "script") +
        ("language model*" | "NLP" | "natural language processing" | "LM" | "large language model*")
      }
    }
  \end{minipage}
  \caption{
    \textbf{Replication: Search Keywords.}
    Using the Semantic Scholar Bulk API, we use the following search keywords to obtain a broad set of literature spanning multilinguality and edge NLP.
  }
  \label{fig:search_keywords}
\end{figure*}

\begin{figure*}[ht]
  \small
  \centering
  \begin{minipage}{0.95\textwidth}
    \promptbox[\textbf{LM Annotation Prompt:} Annotate papers on different dimensions based on their title and abstract.]{
      You are an expert research paper annotator specializing in NLP and machine learning.
      Given the title and abstract of a research paper, classify it according to several dimensions
      relevant to multilingual and efficient NLP for edge deployment.
      Your responses must strictly adhere to the specified response schema without adding any additional commentary.\\\\
      Title: \texttt{\{title\}}\\
      Abstract: \texttt{\{abstract\}}\\\\
      Please classify this paper according to the following dimensions:\\
      1. The single primary pipeline stage the paper addresses (Data Collection, Pretraining, Post-training, Inference, Evaluation, or Full-Stack if it spans 4+ stages).\\
      2. Primary topic(s) or techniques used in the paper.\\
      3. Primary subject area of the paper based on ACL 2025 Subject Areas.\\
      4. Modality of the work (e.g., text, speech, multimodal).\\
      5. Languages studied or supported (if applicable). Use ISO 639-1 codes where possible (e.g., en, fr, de, es). Use "multilingual" if >10 languages.\\
      6. Names of any models released by the authors (empty list if none).\\
      7. Parameter sizes of the models released in billions (empty list if none or not specified).\\
      8. Is this paper primarily about efficiency, multilinguality, both, or neither?\\
      9. What type of contribution does this paper make (Method, Technique, Evaluation, Survey, Resource, Analysis)?\\
      10. Is this paper relevant in the context of multilingual NLP for edge devices? Score from 1 to 5.\\
      11. State your reason for the relevance score.\\
      12. Extract a list of free-form keywords that capture the paper's key concepts, methods, datasets, or findings.\\

    }
  \end{minipage}
  \caption{
    \textbf{Replication: LM Annotation Prompt.}
    Using GPT 4.1 Mini (gpt-4.1-mini-2025-04-14), we annotate several
    papers based on their title and abstract across several dimensions.
    We use the \texttt{outlines} library for structured output generation \citep{willard2023efficientguidedgenerationlarge}.
  }
  \label{fig:appendix_annotation_prompt}
\end{figure*}


\input{tables/lm_pipeline_edge_lm_systems.tex}

\end{document}

%% file: macros.tex
\definecolor{cblue}{HTML}{8EE8D8}
\definecolor{clightblue}{HTML}{D1F9F1}
\definecolor{cwarmblue}{HTML}{00BDB6}
\definecolor{cdarkblue}{HTML}{133844}

\definecolor{clightcrest}{HTML}{FFE2C8}
\definecolor{cwarmcrest}{HTML}{FFC392}
\definecolor{ccrest}{HTML}{FD8153}
\definecolor{cdarkcrest}{HTML}{DD3025}

\definecolor{clightcherry}{HTML}{F2CAD8}
\definecolor{cwarmcherry}{HTML}{E18AAC}
\definecolor{ccherry}{HTML}{CD3572}
\definecolor{cdarkcherry}{HTML}{911449}

\definecolor{clightpurple}{HTML}{F2ECF8}
\definecolor{cwarmpurple}{HTML}{D1B7EB}
\definecolor{cpurple}{HTML}{A368DF}
\definecolor{cdarkpurple}{HTML}{681FB1}

\definecolor{clightindigo}{HTML}{EBEDFB}
\definecolor{cwarmindigo}{HTML}{B0B9F1}
\definecolor{cindigo}{HTML}{5366E0}
\definecolor{cdarkindigo}{HTML}{29347A}

\definecolor{clightgreen}{HTML}{DFF2EA}
\definecolor{cwarmgreen}{HTML}{AFDFCB}
\definecolor{cgreen}{HTML}{4DB78C}
\definecolor{cdarkgreen}{HTML}{13553A}

\definecolor{cslate1}{HTML}{ECEEF1}
\definecolor{cslate2}{HTML}{B5BDC8}
\definecolor{cslate3}{HTML}{546072}
\definecolor{cslate4}{HTML}{232830}

\definecolor{cheritageblue}{HTML}{85B09A}
\definecolor{cjudgeyellow}{HTML}{FFB81C}

\newcommand{\numpapers}{232}
\newcommand{\numdeploymentpapers}{36}

\newcommand{\iconmemory}{\raisebox{-1.5pt}{\includegraphics[height=1.05em]{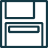}}\xspace}
\newcommand{\iconcompute}{\raisebox{-1.5pt}{\includegraphics[height=1.05em]{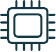}}\xspace}
\newcommand{\iconenergy}{\raisebox{-1.5pt}{\includegraphics[height=1.05em]{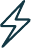}}\xspace}
\newcommand{\icondata}{\raisebox{-1.5pt}{\includegraphics[height=1.05em]{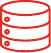}}\xspace}
\newcommand{\iconrepresentation}{\raisebox{-1.5pt}{\includegraphics[height=1.05em]{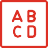}}\xspace}
\newcommand{\iconalignment}{\raisebox{-1.5pt}{\includegraphics[height=1.05em]{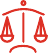}}\xspace}

\newcommand{\promptbox}[2][]{%
  \begin{tcolorbox}[
      rounded corners,
      colback=white,
      colframe=cblue,
      boxrule=1pt,
      toptitle=1mm,              
      bottomtitle=1mm,           
      left=10pt,                 
      right=10pt,                
      top=5pt,                   
      bottom=5pt,                
      title=#1,
      colbacktitle=cblue,
      coltitle=black,
      boxrule=0.5mm,             
      boxsep=5pt,                
    ]
    #2
  \end{tcolorbox}%
}

%% file: authors.tex
\author{
    Lester James V. Miranda$\textsuperscript{1}$\quad
    Songbo Hu$\textsuperscript{1}$\quad
    Roi Reichart$\textsuperscript{2}$\quad
    Anna Korhonen$\textsuperscript{1}$
    \\
    \ \\
    $\textsuperscript{1}$Language Technology Lab, University of Cambridge, United Kingdom
    \\
    $\textsuperscript{2}$Technion -- Israel Institute of Technology, Israel
    \\
    \texttt{\{ljvm2,sh2091,alk23\}@cam.ac.uk}
    \\
    \texttt{roiri@technion.ac.il}
}

%% file: figures/taxonomy.tex
\scalebox{0.68}{%
  \begin{forest}
    forked edges,
    for tree={
    grow'=0,
    draw,
    rounded corners,
    minimum height=1em,
    minimum width=5em,
    node options={align=center},
    where level>=2{node options={align=left}}{},
    where level=1{text width=6em}{},
    where level=2{text width=14em}{},
    where level=3{text width=24em}{},
    anchor=west,
    parent anchor=east,
    child anchor=west,
    l sep=1.5em,
    s sep=0.3em,
    tier/.wrap pgfmath arg={tier #1}{level()},
    edge={-latex},
    },
  [{\rotatebox{90}{\textsc{Multilingual Edge LLMs}}}, fill=white, minimum height=1.5em, minimum width=0em, name=tier1
    [\textbf{Data Collection} (\S\ref{sec:data_collection}), fill=clightgreen, for descendants={fill=clightgreen}, name=tier2
      [Limited capacity \iconmemory{} constrains language coverage and robustness to noise \icondata{}, name=tier3
        [\textbf{Data curation and filtering}: \citet{lignos-etal-2022-toward,ali-etal-2025-judging,shen2025dcad,langlais2026common,datologyai2026uberwebinsightsmultilingualcuration}]
        [\textbf{Synthetic data generation}: \citet{pombal2025mprometheus,mora2025artaskingmultilingualprompt,anugraha2026mr3multilingualrubricagnosticreward,chitale2026updeshsynthesizinggroundedinstruction}]
        [\textbf{Vocabulary sharing}: \citet{yuan-etal-2024-vocabulary,li2025rethinking}, name=tier4]
        [\textbf{Noisy labels}: \citet{jin-etal-2021-instance,wang-etal-2023-noise,wu-etal-2023-noisywikihow}]
      ]
    ]
    [\textbf{Pretraining} (\S\ref{sec:pretraining}), fill=cjudgeyellow!25, for descendants={fill=cjudgeyellow!25}
      [Multilingual performance scales with model size \iconrepresentation{}{,} which is limited in edge devices \iconmemory{}
        [\textbf{Tokenizer design}: \citet{petrov2023language,ali-etal-2024-tokenizer,yu-etal-2023-megabyte,minixhofer2026bolmobyteifyinggenerationlanguage}]
        [\textbf{Continual pretraining}: \citet{kim2024efficienteffectivevocabularyexpansion,cui2024efficienteffectivetextencoding,elhady-etal-2025-emergent,li2025rethinking}]
      ]
      [Multilinguality adds complexity beyond task and domain \iconrepresentation{}{,} straining edge compute \iconcompute{}
        [\textbf{Parameter-efficient methods}: \citet{pfeiffer-etal-2020-mad,pfeiffer-etal-2021-unks,pfeiffer-etal-2022-lifting,parovic-etal-2023-cross,whitehouse-etal-2024-low,dong-etal-2025-mlas,zhan-etal-2024-prefix}]
        [\textbf{Language experts}: \citet{blevins-etal-2024-breaking,li-etal-2025-group,zhao2024sparse,bandarkar2026multilingualroutingmixtureofexperts}]
      ]
    ]
    [\textbf{Post-training} (\S\ref{sec:post_training}), fill=clightpurple, for descendants={fill=clightpurple}
      [Base models for finetuning \iconalignment{} tend to be too large to fit into device memory \iconmemory{}
        [\textbf{Model Compression}: \citet{frantar2023optq,lin-etal-2025-awq,zeng-etal-2024-multilingual,kim-etal-2024-pruning,chimoto-etal-2026-calibrating,kurz-etal-2026-limitations}]
        [\textbf{Knowledge Distillation}: \citet{agarwal2024onpolicydistillationlanguagemodels,gholami-etal-2024-gold,minixhofer2025universal,boizard2025towards}]
      ]
      [Adding/removing capabilities \iconalignment{} requires expensive retraining \iconcompute{}
        [\textbf{Model Merging}: \citet{yadav-etal-2023-tiesmerging,huang-etal-2024-chat,aakanksha2024mix,tao-etal-2024-unlocking}]
        [\textbf{Machine Unlearning}: \citet{liu-etal-2025-rethinking,choi-etal-2024-cross,hwang2025uncoveringpotentialrisksunlearning,farashah2026multilingualamnesiatransferabilityunlearning}]
        [\textbf{Federated Learning}: \citet{weller-etal-2022-pretrained,zhao2024breaking,li-etal-2025-multilingual-federated}]
      ]
    ]
    [\textbf{Inference} (\S\ref{sec:inference}), fill=clightcherry, for descendants={fill=clightcherry}
      [Cost of inference \iconenergy{} is not uniform across languages \iconrepresentation{}
        [\textbf{Prompt Compression}: \citet{li-etal-2025-prompt,jiang-etal-2023-llmlingua,pan-etal-2024-llmlingua,li-etal-2023-compressing,chuang-etal-2024-learning}]
        [\textbf{Speculative Decoding}: \citet{leviathan-etal-2023-fast,yi-etal-2024-towards,timor2025accelerating}]
        [\textbf{Inference-time Adapt.}: \citet{kaplan2025from,zheng2025broken,goddard2025trainingfreetokenizertransplantationorthogonal}]
      ]
    ]
    [\textbf{Evaluation} (\S\ref{sec:evaluation}), fill=cslate1, for descendants={fill=cslate1}
      [Evaluating multilingual LMs \icondata{} for a specific task or language is expensive \iconcompute{}
        [\textbf{Lite Benchmarks}: \citet{singh-etal-2025-global,ojo-etal-2025-afrobench,micallef-borg-2025-melabenchv1}]
        [\textbf{Instance Selection}: \citet{hofmann2025fluid,zouhar-etal-2025-how,li2025active}]
        [\textbf{Dynamic Evaluation}: \citet{kim2025benchhub,shashidhar2025yourbench}]
      ]
    ]
  ]
  \end{forest}%
}

%% file: tables/lm_pipeline_edge_lm_systems.tex
\begin{table*}[ht]
    \small
    \centering
    \resizebox{\textwidth}{!}{%
        \begin{tabular}{p{3cm} p{2.5cm} p{2.5cm} p{2.5cm} p{2.5cm} p{2.5cm}}
            \toprule
    \textbf{System} & \textbf{Data Coll.} & \textbf{Pretraining} & \textbf{Post-Training} & \textbf{Inference} & \textbf{Evaluation} \\
    \midrule
    \textbf{\citet{anikina-2023-towards}}
        Multilingual dialogue model for disaster response optimized for small memory capacity.
        & Data curation of dialogues during robot-assisted disaster response training sessions.
        & Parameter-efficient finetuning using adapters trained on top of the BERT encoder model.
        & --
        & --
        & Evaluation on slot-tagging and dialogue classification tasks. Performed model size comparisons vs. vanilla BERT.
    \\
    \textbf{\citet{singh2024farmerchatscalingaipoweredagricultural}}
        Telegram-based chatbot to help farmers in Kenya, India, Ethiopia, and Nigeria.
        & Data curation of a knowledge base consisting of agricultural practices, product catalogs, and papers.
        & --
        & Start with a base model with reasoning and tool-use, integrated into a RAG system.
        & Inference is chat-centered, and deployed as an API-accessible application using Telegram.
        & Evaluation consists of testing on a held-out test set and user interviews.
    \\
    \textbf{\citetalias{omnilingualasrteam2025omnilingualasropensourcemultilingual} \citeyearpar{omnilingualasrteam2025omnilingualasropensourcemultilingual}}
    Industry-focused with deployments to health practitioners in Nigeria to transcribe Hausa.
        & Data curation from existing ASR data and collection of new data from partners from Mozilla Foundation's Common Voice.
        & Continual pretraining of existing speech encoders on four sizes: 300M, 1B, 3B, and 7B. 
        & Finetuning on low-resource languages and creating a template conditioned on language codes for future SFT.
        & --  
        & Evaluation on 1600+ languages across different dimensions (family, resource availability, etc.)
    \\
    \textbf{\citet{ramjee-etal-2025-ashabot}}
        WhatsApp-based chatbot to address the information needs of community-health workers in rural India.
        & Data curation of a knowledge base in partnership with a research collective (Khushi Baby). Synthetic data for prompt optimization.
        & --
        & Start with a strong base model (GPT-4) where the context is improved continuously using past dialogues and expert inputs.
        & Inference is chat-centered and deployed as an API via WhatsApp.
        & Evaluation is qualitative, centering on user interviews. Evaluation on expert-annotated golden test set.
    \\
    \textbf{\citet{ye-etal-2025-federated}}
        Federated few-shot hate speech detection for marginalized communities in low-resource languages.
        & Prompt-based collection from social media, forums, and news in Afrikaans, Ukrainian, Russian, and Korean. Released as the REACT dataset.
        & Used multilingual BERT (179M) and multilingual DistilBERT (135M) as compact backbone models.
        & Few-shot federated finetuning with personalization via FedPer (private final layers) and adapters.
        & Simulated federated deployment using the Flower framework with one server and four client instances.
        & Macro-F1 across zero-shot and few-shot settings, compared against Perspective API and single-target finetuning baselines.
    \\
    \textbf{\citet{rutunda-etal-2026-large}}
        LLM-based clinical decision support for community health workers in Rwanda (English and Kinyarwanda).
        & 5,609 clinical vignettes from 101 CHWs across 4 Rwandan districts, collected as voice recordings via a custom mobile app (Mbaza), transcribed and translated by linguists.
        & --
        & Prompt engineering with a system prompt tailored to the Rwandan CHW context. Meditron-70B self-hosted on 2 A100 GPUs.
        & API-based inference for commercial models. Meditron-70B hosted on a private cloud instance (GCP). Data collection via custom Mbaza mobile app.
        & 506 Q\&A pairs evaluated by expert clinicians on 11 metrics (adapted Med-PaLM-2 framework) using a 5-point Likert scale, in both English and Kinyarwanda.
    \\
    \bottomrule
        \end{tabular}}
    \caption{
        \textbf{Complementary: Methods used by edge LM systems across the LM pipeline.}
        We show a representative sample of the \numdeploymentpapers{} deployed systems from the full \numpapers{} surveyed papers, and describe which methods (\S\ref{sec:survey}) they used across the LM development pipeline (\S\ref{sec:lm_pipeline}).
        A ``--'' indicates that the step was not applicable or not discussed in the paper.
    }
    \label{tab:lm_pipeline_edge_lm_systems}
\end{table*}